\documentclass[10pt,twocolumn,letterpaper]{article}

\usepackage[pagenumbers]{cvpr} 

\usepackage{graphicx}
\usepackage{amsmath}
\usepackage{adjustbox}
\usepackage{amssymb}
\usepackage{booktabs}
\usepackage{verbatim}
\usepackage{tablefootnote}
\usepackage{stfloats}

\usepackage[normalem]{ulem}
\usepackage{ifthen}
\usepackage{enumitem}
\usepackage{multicol}
\usepackage{multirow}
\newcommand{\anna}[2][]{%
    \ifthenelse{ \equal{#1}{} }
        {\textcolor{blue}{(Anna) #2}}
        {\textcolor{blue}{(Anna) \sout{#1\xspace}#2}}
}
\newcommand{\George}[2][]{%
    \ifthenelse{ \equal{#1}{} }
        {\textcolor{magenta}{(George) #2}}
        {\textcolor{magenta}{(George) \sout{#1\xspace}#2}}
}

\newcommand{\todo}[2][]{%
    \ifthenelse{ \equal{#1}{} }
        {\textcolor{red}{TODO: #2}}
        {\textcolor{red}{TODO: \sout{#1\xspace}#2}}
}

\usepackage[pagebackref,breaklinks,colorlinks]{hyperref}

\usepackage[capitalize]{cleveref}
\crefname{section}{Sec.}{Secs.}
\Crefname{section}{Section}{Sections}
\Crefname{table}{Table}{Tables}
\crefname{table}{Tab.}{Tabs.}

\def\cvprPaperID{*****} 
\def\confName{CVPR}
\def\confYear{2024}

\begin{document}

\title{Semantic Style Transfer for Enhancing Animal Facial Landmark Detection}

\date{} 

\author{Anadil Hussein \qquad Anna Zamansky \qquad George Martvel \\
\small 
University of Haifa, Israel
}

\maketitle

\noindent

\bigbreak

\begin{abstract}
    Neural Style Transfer (NST) is a technique for applying the visual characteristics of one image onto another while preserving structural content. Traditionally used for artistic transformations, NST has recently been adapted, e.g., for domain adaptation and data augmentation. This study investigates the use of this technique for enhancing animal facial landmark detectors training. As a case study, we use a recently introduced Ensemble Landmark Detector for 48 anatomical cat facial landmarks and the CatFLW dataset it was trained on, making three main contributions. First, we demonstrate that applying style transfer to cropped facial images rather than full-body images enhances structural consistency, improving the quality of generated images. Secondly, replacing training images with style-transferred versions raised challenges of annotation misalignment, but Supervised Style Transfer (SST)~--- which selects style sources based on landmark accuracy - retained up to 98\% of baseline accuracy. Finally, augmenting the dataset with style-transferred images further improved robustness, outperforming traditional augmentation methods. These findings establish semantic style transfer as an effective augmentation strategy for enhancing the performance of facial landmark detection models for animals and beyond. While this study focuses on cat facial landmarks, the proposed method can be generalized to other species and landmark detection models.
\end{abstract}
\section{Introduction}
\label{sec:intro}
Neural Style Transfer (NST) is a research field studying the applications of neural networks to render images in different styles by retaining the structure of the original content while adopting the visual elements of a chosen style~\cite{jing2019neural, singh2021neural}. NST has numerous applications in art generation, photo, and video editing and many more~\cite{v1Gayts2015ArtisticStyle, v1JohnsonSuperResoluation2016, v1ParkStyle-Attentional2018}. Recent advancements in Vision Transformers (ViTs)~\cite{vit} have introduced a new paradigm in style transfer, applying self-attention mechanisms to capture long-range morphological dependencies in an image. Splice ViT~\cite{tumanyan2022splicing} introduced a novel approach to semantic visual appearance transfer, ensuring that style transformations maintain spatial structure. Unlike GAN-based models~\cite{goodfellow2014generative} that rely on adversarial training, Splice ViT leverages pre-trained Vision Transformers to encode content and style separately, allowing for more controlled transformations.

This study explores the use of NST algorithms in a novel domain~--- animal affective computing. Automated analysis of facial expressions is a crucial challenge in this emerging field~\cite{Broome2022}. One of the most promising approaches in this context is facial landmarks, which are well-studied for humans~\cite{wu2018look, belh13lfpw, Le12helen} and are now also being adopted for animals~\cite{Martvel2024Detection, pessanha2022facial, hewitt2019pose, martvel2024dogflw}. 

Facial landmarks (keypoints/fiducial points) detection~\cite{wu2019facial} is extensively studied for human automated facial analysis. Landmark locations have been shown to provide crucial insights in the context of face alignment, feature extraction, facial expression recognition, head pose estimation, eye gaze tracking, and other tasks~\cite{gazeestimation, MAlEidan2020DeepLearningBasedMF, MALEK2021406}. They have also been used for automated facial movement recognition systems~\cite{FAU, AUIP}. An important advantage of landmark-based approaches is their ease of application to video data, producing time series of multiple coordinates, which can then be analyzed~\cite{zhan2021key}, classified~\cite{sehara2021real, ferres2022predicting, liu2022murine}, or processed for different purposes~\cite{hardin2022using, suryanto2022using}. 

Landmark-based approaches are just beginning to be explored in the domain of animal behavior, primarily for body landmarks (e.g.,~\cite{labuguen2019primate, wiltshire2023deepwild, jeon2023deep}), while facial analysis still remains relatively underexplored. One significant challenge arises from the variety of textures, shapes, and morphological structures found across different breeds and species. This is especially true for domesticated animals, such as farm and companion ones, which exhibit significant variability in appearance and morphology ~\cite{vila1999phylogenetic}. 
Another challenge in this field is the lack of large, high-quality datasets. Typically, in the human domain, datasets with facial landmarks consist of thousands of images with dozens of landmarks. The animal domain, on the other hand, severely needs landmark-related datasets and benchmarks, as highlighted in Broomé et al.~\cite{broome2023going}. These are just beginning to be developed for species such as cats~\cite{sun20cafm}, dogs~\cite{liu2012dog}, horses~\cite{pessanha2022facial}, cattle~\cite{coffman2024cattleface}, and sheep~\cite{hewitt2019pose}; however, in most cases, the limited number of instances in the training data, the small number of landmarks, as well as lack of justification for their placement in terms of facial muscles, makes these tools inadequate for capturing the subtle facial changes necessary for emotion or pain recognition. 

Martvel et al.~\cite{martvel2024automated} recently addressed this gap for cat and dog facial analysis by introducing datasets and detector models for 48 and 46 anatomy-based facial landmarks for cats and dogs, respectively~\cite{Martvel2024Detection, martvel2024dogflw}. The detector has been utilized in a number of studies, performing well in tasks requiring subtle facial analysis, such as breed, cephalic type, and pain recognition in cats~\cite{martvel2024automatedA, martvel2024automatedD}. However, data scarcity still remains an issue on the way to better performance in the task of landmark detection. 

To address the bottlenecks of data scarcity, data augmentation techniques such as random cropping, flipping, rotation, and brightness adjustments \cite{dataAugmentation} are commonly used. In this study, we explore a more sophisticated augmentation using NST, while prior work has explored semi-supervised style transfer for boosting facial landmark detection in human faces ~\cite{styleGAN}, we apply style transfer on animals for the first time in this context, investigating both supervised and semi-supervised style transfer approaches, handling the morphological variability of animal species. In this study, we explore Neural Style Transfer as a data augmentation technique that stylistically transforms animal facial images while preserving key morphological structures crucial for landmark detection.

\section{Methodology}
\label{sec:Methodology}

\subsection{Model}
For style transfer, we employ Splice ViT \cite{Tumanyan2022SpliceViT}, a Vision Transformer-based approach that ensures semantic consistency during style transformations. Unlike traditional NST methods, which can disrupt spatial alignment, Splice ViT separates content and style representations, preserving facial structures critical for landmark detection.
One of the challenges in applying style transfer to landmark-based models is annotation misalignment—where stylistic transformations alter the image while retaining the original landmark coordinates. To mitigate this, we incorporate Supervised Style Transfer (SST), selecting style reference images with minimal landmark displacement, ensuring that transformations do not compromise spatial accuracy.

\subsection{Dataset and Preprocessing}
This study utilizes the Cat Facial Landmarks in the Wild (CatFLW) dataset \cite{Martvel2023CatFLW}, which consists of 2,091 annotated feline facial images with 48 anatomical facial landmarks. The dataset provides a diverse representation of cat faces, enabling robust training for landmark detection models.
To maintain structural consistency in style transfer, we preprocess the dataset by isolating the cat’s face region from the full image. Prior studies suggest that applying style transfer to full-body images can introduce background artifacts, feature distortions, and misalignment of facial landmarks \cite{Martvel2024Analysis}. To mitigate these issues, we compare two preprocessing strategies:
\begin{enumerate}
    \item Full-Body Style Transfer (FB-ST): Style transfer is applied to the entire image, preserving background context.
    \item Cropped-Face Style Transfer (CF-ST): The cat’s face is extracted before style transfer to minimize distortions.
\end{enumerate}
We evaluated which preprocessing method better preserves facial structure by testing 30 random image pairs using both FB-ST and CF-ST for style transfer. The structural preservation was evaluated using two metrics: the Intersection over Union (IoU) between the original and style-transferred segmentation masks, and the Splice ViT loss function ($L_{\text{splice}}$), which is the total loss used during training. The $L_{\text{splice}}$ loss combines appearance preservation, structural consistency, and identity retention to guide the model in producing realistic style-transferred outputs while maintaining both texture and structure.

\begin{figure}[h!]
            \centering
          \includegraphics[width=0.8\linewidth]{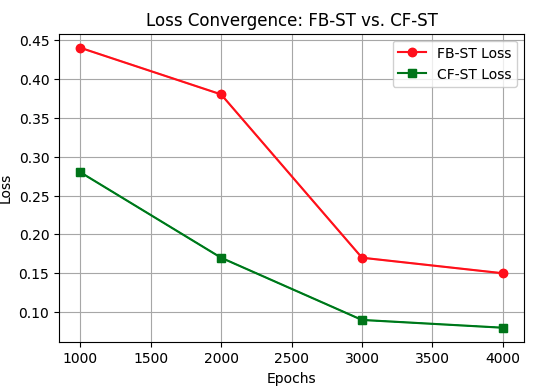}  \vspace{-8pt}
    \caption{ The convergence behavior of the model over epochs.}
    \label{fig:combined}
\end{figure}

The experimental results showed that the CF-ST method consistently outperformed FB-ST across all evaluation metrics. As shown in Figure~\ref{fig:combined}, CF-ST achieved a lower loss at each checkpoint. At 1,000 epochs, the average loss for FB-ST was 0.44, compared to 0.28 for CF-ST. By 4,000 epochs, CF-ST reached a final loss of 0.08, while FB-ST remained at 0.15, indicating that the model converged faster and more stably when operating on cropped faces.

\vspace{-4pt}
\vspace{-4pt}

\begin{figure}
    \centering
    \includegraphics[width=\linewidth]{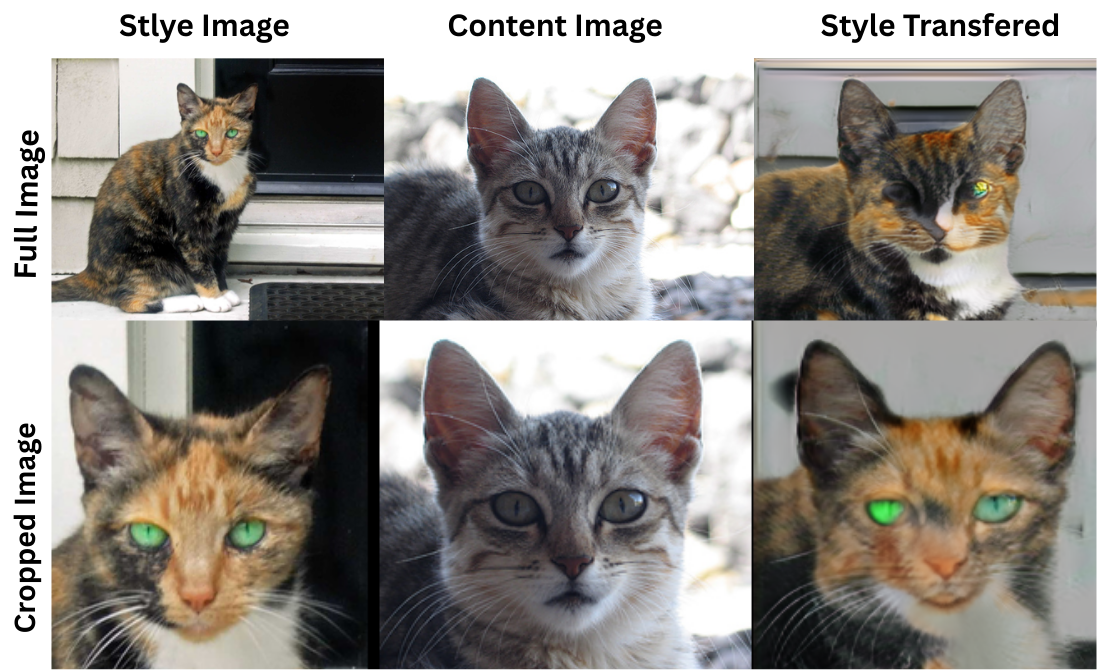}
    \caption{Comparison between Full-Body Style Transfer (FB-ST) and Cropped-Face Style Transfer (CF-ST).}
    \label{fig:Comparison}
\end{figure}

In terms of IoU, CF-ST achieved an average score of 0.8574, compared to 0.4634 for FB-ST. This substantial difference demonstrates that cropping the face before applying style transfer significantly improves structural preservation. Figure~\ref{fig:Comparison} provides examples of how facial features are better maintained in the CF-ST outputs.

Based on the initial experiments, we concluded that cropping the face before applying style transfer leads to more stable, consistent, and structurally accurate outputs. Therefore, all subsequent experiments in this study were conducted using the CF-ST preprocessing approach.

\subsection{Experimental Setup}
A subset of 600 images was randomly selected from the CatFLW dataset and split into 500 training images (\textit{Train}) and 100 test images (\textit{Test}). The subset size was chosen to ensure computational feasibility, as each Splice ViT transformation takes approximately 20 minutes per image pair on an NVIDIA A100 GPU.

All models were trained under identical conditions using Splice ViT with a learning rate of 0.05 and optimized over 4,000 epochs. Performance was evaluated using Normalized Mean Error (NME), Failure Rate (FR, thresholded at NME $>$ 10), and region-specific NME for localized accuracy.

The experiments were structured to evaluate the effect of style transfer across different training scenarios. A baseline model was first trained on original images and evaluated on the original test set. To test generalization, the same model was also evaluated on a style-transferred version of the test set (\textit{TestST}), where each test image was stylized using a random training image.

Next, a model was trained on a fully style-transferred training set (\textit{TrainST}), replacing all original images with their stylized versions. To reduce misalignment caused by uncontrolled styling, we introduced a Supervised Style Transfer (SST) strategy, where style sources were selected based on landmark prediction accuracy (lowest NME). The selection process is defined as:
\vspace{-4pt}
\begin{equation}
\text{SST}(N) = \left\{ I_i \mid \arg\min_N \left( \text{NME}(I_i) \right), I_i \in \text{Train} \right\}
\end{equation}
\begin{equation}
\text{TrainSST}(N) = \left\{ G(I, S) \mid I \in \text{Train}, S \sim \text{SST}(N) \right\}
\end{equation}
where $I$ is a training image, $S$ is a style source randomly selected from the SST($N$) set, and $G(I, S)$ is the style transfer function.

To study style diversity, three configurations were tested:
\begin{itemize}
    \item \textbf{TrainSST(1)} – stylized using a single best-performing style sources.
    \item \textbf{TrainSST(10)} – using one of the top 10 sources.
    \item \textbf{TrainSST(250)} – using one of the top 250 sources.
\end{itemize}
In addition, style-transferred images were also used as augmentation by combining them with the original training data (Train + TrainST, Train + TrainSST(N)). A control experiment was conducted with rotation-based augmentation (Train + TrainRotated), where rotated duplicates of training images were added to assess traditional augmentation.

\section{Results}
\subsection{Training on Style-Transferred Data Alone}
\begin{figure*}[ht]
    \centering
    \begin{minipage}{0.48\linewidth}
        \centering
        \includegraphics[width=\linewidth]{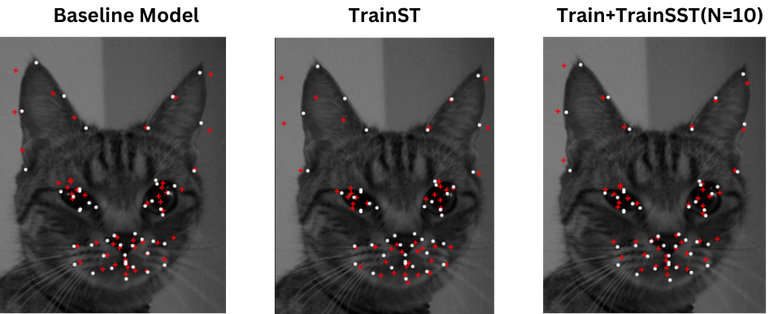}
        \caption{Qualitative comparison of predicted landmarks (white) and ground truth landmarks (red) across three training setups: Baseline model trained on original data (left), model trained on style-transferred data (TrainST), and model trained on a combination of original and supervised style-transferred data (Train+TrainSST(N=10)).}
        \label{fig:qual_landmarks}
    \end{minipage}
    \hfill
    \begin{minipage}{0.48\linewidth}
        \centering
        \includegraphics[width=\linewidth]{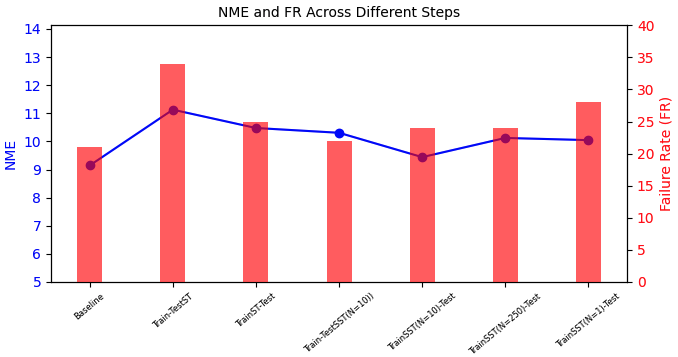}
        \caption{Comparison of Normalized Mean Error (NME) and Failure Rate (FR) across different training strategies.}
        \label{fig:compare_nme_fr}
    \end{minipage}
\end{figure*}

The baseline model (\textbf{Train}-\textbf{Test}) achieved an NME of 9.144, reaching the highest accuracy under original training conditions. When the model was trained entirely on style-transferred images (\textbf{TrainST}-\textbf{Test}), NME increased to 10.477, leading to a 14.6\% worse performance compared to the baseline. Training on SST with the top 10 most confidently predicted images (\textbf{TrainSST(N=10)}-\textbf{Test}) resulted in an NME of 9.441 (compared to 9.144 in the baseline model). This represents a significant reduction in performance loss from unsupervised ST with 14.6\% down to just 3.2\%, demonstrating that controlling style sources helps the model retain crucial facial structure information (see Fig.~\ref{fig:qual_landmarks}). Increasing the number of style sources to 250 images (\textbf{TrainSST(N=250)}-\textbf{Test}) led to an NME of 10.123, while using only one style source (\textbf{TrainSST(N=1)}-\textbf{Test}) resulted in NME of 10.046, maintaining 90.9\% of the baseline accuracy (Fig.~\ref{fig:compare_nme_fr}).

\subsection{Style-Transfer-Based Data Augmentation}
Table~\ref{tab:augmentation_results} shows that all style-transfer-based augmentation methods outperformed the baseline. The best performance was achieved using \textbf{Train + TrainSST (N=1)}, with the lowest NME of 7.638 and FR of 11. Figure~\ref{fig:per-region} further illustrates that these strategies reduced NME across facial regions compared to the baseline. The rotation-based augmentation (Train + TrainRotated) resulted in higher NME (9.829), reinforcing the advantage of style-based semantic augmentation.

\begin{table}[!b]
\centering
\caption{Comparison of training configurations on NME and Failure Rate (FR). Lower values indicate better performance.}
\begin{tabular}{|l|c|c|}
\hline
\textbf{Training Configuration} & \textbf{NME} & \textbf{FR} \\
\hline
Baseline (Original Only)        & 9.144        & 21 \\
TrainST (Random Style)          & 10.451 & 25 \\
TrainSST (N=10)                 & 9.870  & 24 \\
TrainSST (N=250)                & 9.962  & 24 \\
TrainSST (N=1)                  & 10.482 & 28 \\
Train + TrainRotated            & 9.256  & 14 \\
Train + TrainST (Random)        & 7.780        & 11 \\
Train + TrainSST (N=1)          & \textbf{7.638}        & \textbf{11} \\
Train + TrainSST (N=10)         & 7.838        & 14 \\
Train + TrainSST (N=250)        & 8.624        & 13 \\
\hline
\end{tabular}
\label{tab:augmentation_results}
\end{table}

\begin{figure*}[!b]
    \centering
    \includegraphics[width=0.8\linewidth]{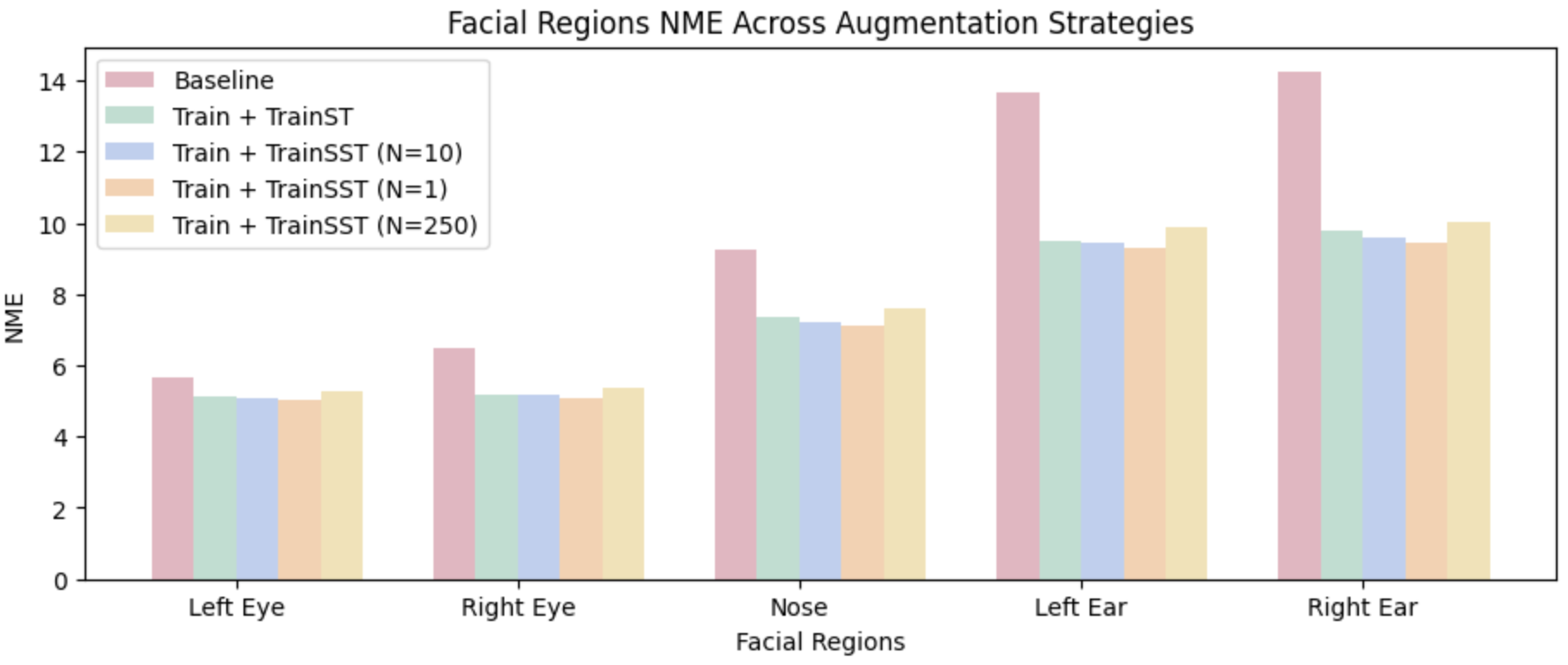}
    \caption{Comparison of Normalized Mean Error (NME) across different facial regions among the augmentation strategies}
    \label{fig:enter-label}
\end{figure*}
\section{Conclusion}
\label{sec:conclusion}
This study investigated the application of Neural Style Transfer (NST) in a novel domain, such as animal facial landmark detection, making three key contributions. First, in terms of preprocessing, our findings demonstrate that cropping the faces—rather than applying transformations to full-body images—greatly enhances structural consistency. This suggests that face cropping is a crucial step when using style transfer in this domain. Secondly, our results indicate that while training on style-transferred images alone degrades performance due to annotation misalignment, supervised selection of style sources effectively mitigates this issue. Furthermore, augmenting original data with curated style-transferred images leads to notable improvements in model accuracy. Moreover, supplementing the original dataset with carefully curated style-transferred images yields significant improvements in model accuracy. These findings suggest that NST can be a valuable tool for enhancing dataset diversity and boosting generalization in animal affective computing tasks such as landmark detection. This positions semantic style transfer as a promising method for generating synthetic data, critical for scaling training sets for large models requiring more diverse data.
\newpage
{\small
\bibliographystyle{plain}
\bibliography{anadilSections/ref}
}

\end{document}